\documentclass[10pt,twocolumn,letterpaper]{article}

\usepackage{cvpr}
\usepackage{times}
\usepackage{epsfig}
\usepackage{graphicx}
\graphicspath{ {../images/} }
\usepackage{amsmath}
\usepackage{amssymb}
\usepackage{import}
\usepackage{comment}
\usepackage{multicol}
\usepackage{multirow}
\usepackage{booktabs}
\usepackage{float}

\usepackage[breaklinks=true,bookmarks=false]{hyperref}

\cvprfinalcopy 

\def\cvprPaperID{6} 

\begin{document}

\title{Top-Down Networks: A coarse-to-fine reimagination of CNNs}

\author{Ioannis Lelekas\\
{\tt\small giannislelekas@gmail.com}
\and
Nergis Tomen\\
{\tt\small N.Tomen@tudelft.nl}
\and
Silvia L. Pintea\\
{\tt\small S.L.Pintea@tudelft.nl}
\and
Jan C. van Gemert\\
{\tt\small J.C.vanGemert@tudelft.nl}
\\
Computer Vision Lab,
Delft University of Technology, NL\\
}

\maketitle

\begin{abstract}
   Biological vision adopts a coarse-to-fine information processing pathway, from initial visual detection and binding of salient features of a visual scene, to the enhanced and preferential processing given relevant stimuli. On the contrary, CNNs employ a fine-to-coarse processing, moving from local, edge-detecting filters to more global ones extracting abstract representations of the input.
   In this paper we reverse the feature extraction part of standard bottom-up architectures and turn them upside-down: We propose top-down networks. Our proposed coarse-to-fine pathway, by blurring higher frequency information and restoring it only at later stages, offers a line of defence against adversarial attacks that introduce high frequency noise. 
   Moreover, since we increase image resolution with depth, the high resolution of the feature map in the final convolutional layer contributes to the explainability of the network's decision making process.
   This favors object-driven decisions over context driven ones, and thus provides better localized class activation maps. This paper offers empirical evidence for the applicability of the top-down resolution processing to various existing architectures on multiple visual tasks.
\end{abstract}

\vspace{-0.25cm}
\section{Introduction}\label{sec:Introduction}

\vspace{-0.15cm}
In human biological vision, perceptual grouping of visual features is based on Gestalt principles, where factors such as proximity, similarity or good continuation of features generate a salient percept~\cite{Wagemans2012}.
Salient objects are rapidly and robustly detected and segregated from the background in what is termed the ``pop-out” effect~\cite{Field1993,Kovacs1993}.
This initial detection and grouping of salient features into a coherent percept, leads to preferential processing by the visual system, described as stimulus-driven attention~\cite{Zhang2012}.
For relevant visual stimuli, the exogenously directed attention is sustained, and results in a more detailed visual evaluation of the object.
This typical pipeline of perception and attention allocation in biological vision represents an efficient, coarse-to-fine processing of information~\cite{Hegde2008}.
In contrast, modern CNNs (Convolutional Neural Networks) do not incorporate this perspective \cite{he2016deep, krizhevsky2012imagenet, simonyan2014very, szegedy2015going}.

\begin{figure}[t]
    \begin{center}
        \includegraphics[width=\linewidth]{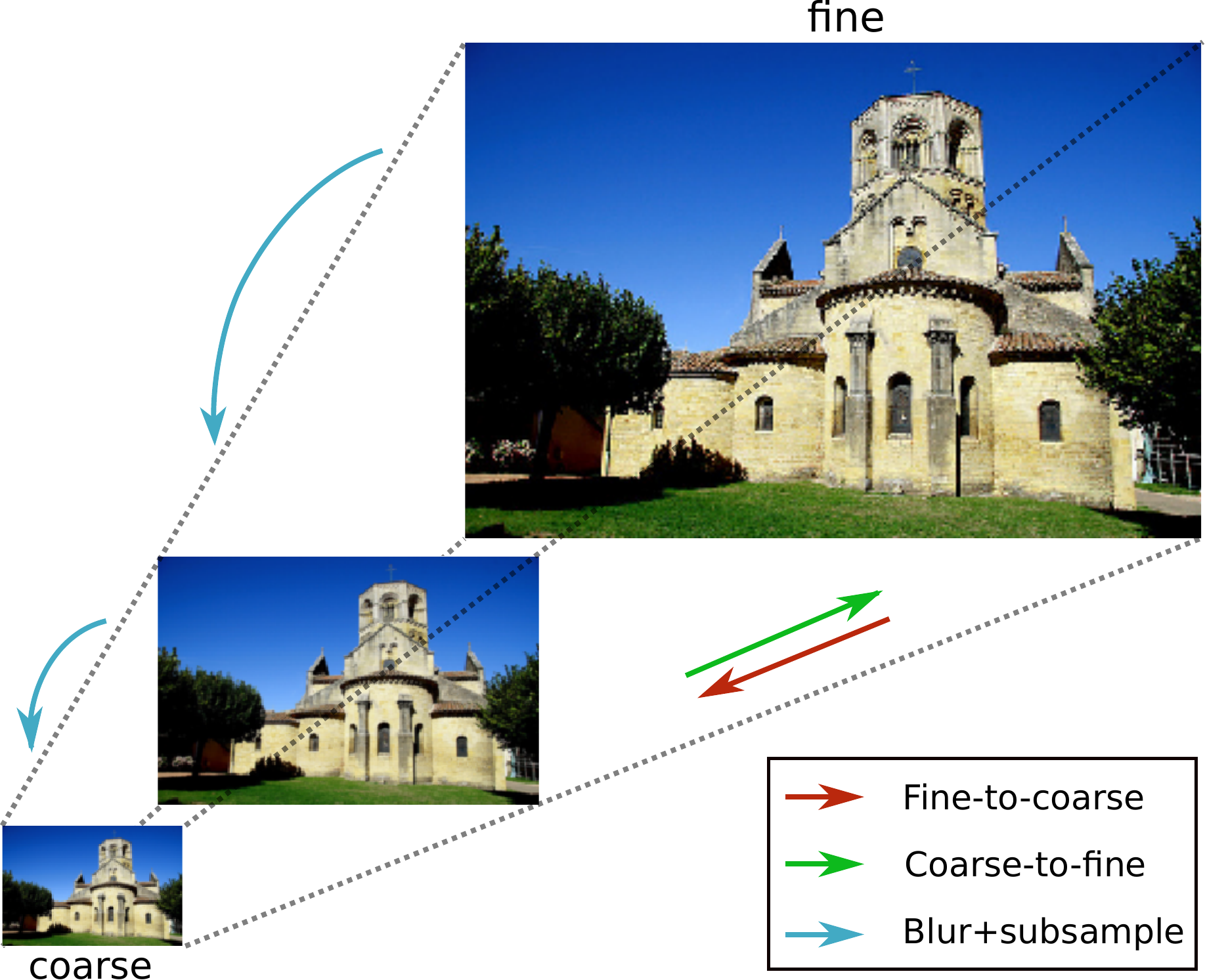}    
    \end{center}
    \vspace{-0.4cm}
    \caption{
    A coarse-to-fine versus fine-to-coarse processing pathway. The conventional fine-to-coarse pathway in a CNN sacrifices localization for semantically richer information. 
    The opposite path, proposed in this paper, starts from the coarsest input and focuses on the context: given the sky, grass and building, it is clearly a landscape scene of a building. 
    Moving to finer representations of the input, the focus shifts to local information. 
    Architectural aspects of the building, and the cross on the top, are now the most informative for classifying the image as a church. 
    Our proposed coarse-to-fine pathway is in line with human biological vision, where detection of global features precedes the detection of local ones, for which further processing of the stimuli is required.}
    \vspace{-0.3cm}
    \label{fig:figure_11}
\end{figure}

Standard CNNs begin with the high resolution input, and propagate information in a fine-to-coarse pathway. Early layers learn to extract local, shareable features, whereas deeper layers learn semantically rich and increasingly invariant representations.
In this paper we propose the reversal of the conventional feature extraction of standard CNNs, as depicted in Figure~\ref{fig:figure_11}.
More specifically, we suggest the adoption of a coarse-to-fine processing of the input, which can be interpreted as gradual focusing of visual attention. The top-down hierarchy first extracts the gist of a scene, starting from a holistic initial representation, and subsequently enhances it with higher frequency information. 

A growing body of literature since the seminal work of \cite{goodfellow2014explaining,szegedy2013intriguing} shows that adversarial perturbations with high-frequency components may cause substantial misclassifications.
Suppressing higher frequencies in the input image, as proposed in our top-down paradigm, can provide a first line of defence.
At the same time, explainability of the decision making process of CNNs has recently emerged as an important research direction~\cite{selvaraju2017grad,zhou2016learning}.
In this context, our coarse-to-fine processing scheme, having feature maps with higher spatial resolution at deeper layers, favors object-driven decisions over context-driven ones, and provides better localized class activation maps. 

We make the following contributions: 
(i) We propose biologically inspired top-down network architectures, obtained by reversing the resolution processing of conventional bottom-up CNNs;
(ii) We analyze various methods of building top-down networks based on bottom-up counterparts as well as the difference in resolution-processing between these models, providing a versatile framework that is directly applicable to existing architectures;
(iii) We compare our proposed model against the baseline on a range of adversarial attacks and demonstrate enhanced robustness against certain types of attacks.
(iv) We find enhanced explainability for our top-down model, with potential for object localization tasks. 
Trained models and source code for our experiments are available online: \href{https://github.com/giannislelekas/topdown}{https://github.com/giannislelekas/topdown}.
\vspace{-0.2cm}
\section{Related work}\label{sec:related_work}
\vspace{-0.2cm}

\vspace{5px}\noindent\textbf{Coarse-to-fine processing.} Coarse-to-fine processing is an integral part of efficient algorithms in computer vision. Iterative image registration~\cite{lucas1981iterative} gradually refines registration from coarser variants of the original images, while in \cite{hu2016efficient} a coarse-to-fine optical flow estimation method is proposed.
Coarse-to-fine face detection is performed by processing increasingly larger edge arrangements in~\cite{fleuret2001coarse}, and coarse-to-fine face alignment using stacked auto-encoders is introduced in~\cite{zhang2014coarse}.
Efficient action recognition is achieved in~\cite{wu2019liteeval} by using coarse and fine features coming from two LSTM (Long Short-Term Memory) modules. 
In~\cite{sahbi2017coarse} coarse-to-fine kernel networks are proposed, where a cascade of kernel networks are used with increasing complexity.
Existing coarse-to-fine methods consider both coarse input resolution, as well as gradually refined processing. Here, we also focus on coarse-to-fine image resolution, however we are the first to do this in a single deep neural network, trained end-to-end, rather than in an ensemble.   

\vspace{5px}\noindent\textbf{Bottom-up and top-down pathways.}
Many approaches exploit high spatial resolution for finer feature localization, which is crucial for semantic segmentation. The U-net \cite{ronneberger2015u} and FPN (Feature Pyramid Networks) \cite{lin2017feature} merge information from bottom-up and top-down pathways, combining semantically rich information of the bottom-up with the fine localization of the top-down stream. 
Similarly, combinations of a high-resolution and a low-resolution branch were proposed for efficient action recognition \cite{fan2019more}, for face hallucination \cite{li2018coarse}, and depth map prediction \cite{eigen2014depth}.
Top-down signals are also used to model neural attention via a backpropagation algorithm \cite{zhang2016topdown}, and to extract informative localization maps for classification tasks in Grad-CAM~\cite{selvaraju2017grad}.
Similarly, we also focus on top-down pathways where we slowly integrate higher levels of detail, however our goal is biologically-inspired resolution processing, rather than feature-map activation analysis.

\vspace{5px}\noindent\textbf{Multi-scale networks.}
Merging and modulating information extracted from multiple scales is vastly popular \cite{honari2016recombinator,ke2017multigrid,yang2019closer,ye2018evenly,xu2014scale}. 
In \cite{ye2018evenly} feature maps are resized by a factor to obtain cascades of multiple resolutions.
Incremental resolution changes during GAN (Generative Adversarial Network) training are proposed in 
\cite{karras2017progressive}.
Convolutional weight sharing over multiple scales is proposed in \cite{aich2020multi,yang2019closer}.
Similarly \cite{fan2019scale} performs convolutions over multiple scales in combination with residual connections.
In \cite{ke2017multigrid} convolutions are performed over a grid of scales, thus combining information from multiple scales in one response, and~\cite{sosnovik2019scale} combines responses over multiples scales, where filters are defined using 2D Hermite polynomials with a Gaussian envelope. 
Spatial pyramid pooling is proposed in \cite{he2015spatial} for aggregating information at multiple scales.
In this work, we also extract multi-resolution feature maps, in order to start processing from the lowest image scale and gradually restore high frequency information at deeper layers.

\begin{figure*}[h!]
	\includegraphics[width=\linewidth]{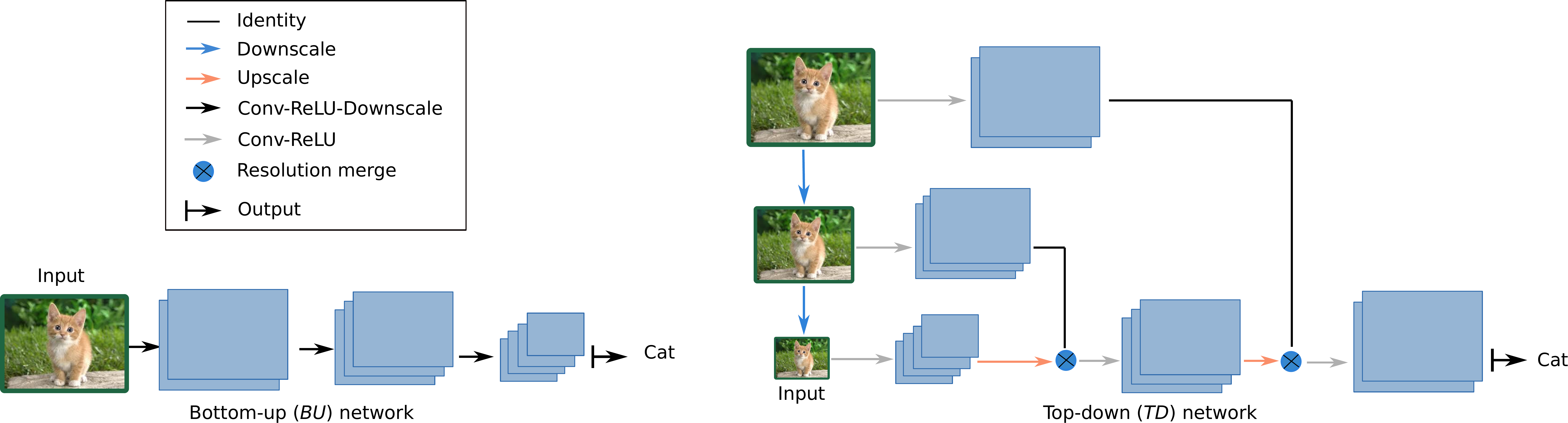}
	\caption{\textbf{Left:} The bottom-up ($BU$) baseline network.
	Feature maps decrease in spatial resolution with network depth.
	\textbf{Right:} The proposed top-down ($TD$) network. 
	The $TD$ model reverses the feature extraction pathway of the baseline network. It employs three inputs from highest to lowest scale, starts processing from the lowest resolution and progressively adds high resolution information.} 
	\label{fig:idea}
	\vspace{-0.3cm}
\end{figure*}

\vspace{5px}\noindent\textbf{Beneficial effects of blurring.}
Suppressing high frequency information by blurring the input can lead to enhanced robustness~\cite{wang2019high,zhang2019adversarial}. 
Models trained on blurred inputs exhibit increased robustness to distributional shift \cite{jo2017measuring}. 
The work in \cite{geirhos2018imagenet} reveals the bias of CNNs towards texture, and analyzes the effect of blurring distortions on the proposed Stylized-ImageNet dataset. 
Anti-aliasing by blurring before downsampling contributes to preserving shift invariance in CNNs \cite{zhang2019making}.
By using Gaussian kernels with learnable variance, \cite{shelhamer2019blurring} adapts the receptive field size. 
Rather than changing the receptive field size, works such as \cite{liang2017detecting,li2017adversarial,raju2019blurnet} use spatial smoothing for 
improved resistance to adversarial attacks. 
Similarly, we also rely on Gaussian blurring before downsampling the feature maps to avoid aliasing effects, and as a consequence we observe improved robustness to adversarial attacks.   

\vspace{-0.15cm}
\section{Top-down networks}\label{sec:method}
\vspace{-0.05cm}

Top-down ($TD$) networks mirror the baseline bottom-up ($BU$) networks, and reverse their feature extraction pathway.
Information flows in the opposite direction, moving from lower to higher resolution feature maps. 
The initial input of the network corresponds to the minimum spatial resolution occurring in the $BU$ baseline network. 
Downscaling operations are replaced by upscaling, leading to the coarse-to-fine information flow. By  upscaling, the network can merely ``hallucinate" higher resolution features.
To restore the high frequency information, we use resolution merges, which combine the hallucinated features with higher frequency inputs, after each upscaling operation.
Figure~\ref{fig:idea} depicts the difference between the $BU$ architecture and our proposed $TD$ architecture.

\subsection{Input and feature map resizing}
To avoid artifacts hampering the performance of the network \cite{zhang2019making}, we blur the inputs before downsampling.
For the upsampling operation we use interpolation followed by convolution. 
We have experimented with both nearest neighbour and bilinear interpolation, and have noticed improved robustness against adversarial attacks for nearest neighbor interpolation. 
We have also considered the use of transpose convolutions, however we did not adopt these due to detrimental checkerboard artifacts.

\subsection{Merging low and high resolution}
\begin{figure}[b]
	\begin{center}
	    \includegraphics[width=0.9\linewidth]{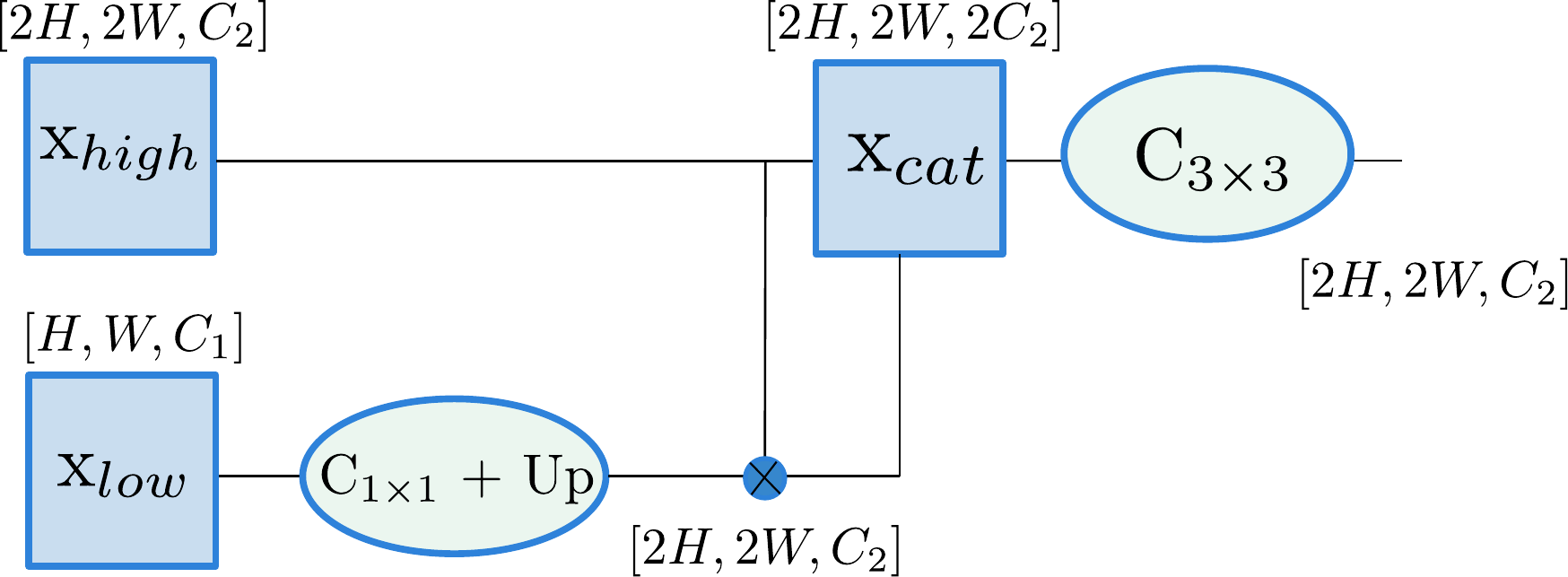}    
	\end{center}
	\caption{Merging low and high-frequency feature maps: we use a $1\times1$ convolution followed by an element-wise addition; this information is concatenated with the high-resolution input and followed by a $3\times3$ convolution that expands the receptive field size.}
	\label{fig:merge_high}
\end{figure}

Figure~\ref{fig:merge_high} depicts the considered method for merging the high resolution input with the low resolution information. 
We first upsample the low resolution input via a $1\times 1$ convolution and use an element-wise addition with the high-resolution branch. 
This information is then concatenated with the original high resolution information on the channel dimension. 
We subsequently use a $3\times3$ convolution to expand the receptive field of the filters. 
The proposed merging of information slightly increases the number of parameters, while being effective in practice.

\vspace{5px}\noindent\textbf{ERF (effective receptive field) size computation.} Neurons in each layer $i$ of a typical bottom-up network has a single ERF size $r_{i}$ determined by the kernel size $k_{i}$ and the cumulative stride $m_{i}$ (given stride $s_{i}$ at layer $i$).
\vspace{-0.15cm}
\begin{alignat}{1}
    r_{i} &= r_{i-1} + (k_{i} - 1) \; m_{i-1} \nonumber \\ 
    m_{i} &= m_{i-1} \cdot s{i}
\label{eq:res}
\end{alignat}
\vspace{-0.5cm}

\noindent Assuming only $3\times3$ convolutions with stride 1, the example $BU$ architecture in Figure~\ref{fig:idea} will have an ERF size of 3 pixels, and 18 pixels in each direction after the first and final convolutional layers, respectively. In contrast, for the $TD$ network, considering a Gaussian blurring window of width $6 \sigma$, the lowest resolution branch will already have an ERF size of $12 \sigma \!+\! 2$ at the input level and of $12 \sigma \!+\! 10$ after the first convolutional layer (comparable to the final layer of a $BU$ network already with $\sigma=2/3$ pixels). Furthermore, in contrast to $BU$, output from neurons with varying ERFs are propagated through the merging points. To get a lower bound on the $TD$ ERF sizes, we consider resolution merging methods which do not provide RF enlargement (e.g. as depicted in fig.~\ref{fig:merge_high}, but without the $3\times3$ convolution at the end).
Thus, at the final merging point of the $TD$ architecture, ERF sizes of 3 pixels and $12 \sigma \!+\! 14$ pixels are merged together. 
In conclusion, already from the first layer, $TD$ has the ERF size that the $BU$ only obtains at the last layer.

\subsection{Filter arrangement}\label{sec:filters_arrangement}
Feature extraction pathway of the $TD$ network reverses the $BU$: information propagates from lower to higher spatial dimensions in a $TD$ network, while the number of filters shrinks with increasing depth. 
The choice of expanding the number of filters at deeper layers in the $BU$ network is efficiency-oriented. As the feature map resolution decreases, the number of channels increases, retaining the computational complexity roughly fixed per layer. 
Typically, in standard architectures the filters are doubled every time dimensions are halved \cite{he2016deep,simonyan2014very}. 

In our method we consider three options for deciding the number of filters per layer: the $TD$ model which is exactly the opposite of the $BU$ in that the number of channels are reduced with depth; the \emph{uniform} model ($TD_{uni}$) where the layers have a uniform number of filters; and the \emph{reversed} model ($TD_{rev}$) which follows the $BU$ filter arrangement, with channel dimension widened with depth.
\section{Experiments}\label{sec:experiments}
In \textbf{Exp 1} we evaluate the three different filter arrangement options proposed for the
top-down model.
We compare these model variations with the bottom-up baseline on the MNIST, Fashion-MNIST and CIFAR10 classification tasks.
In \textbf{Exp 2} we evaluate the robustness of our proposed model against various adversarial attacks applied on the same datasets. 
Finally, in \textbf{Exp 3} we illustrate the explainability capabilities of our top-down model when compared to the bottom-up, and demonstrate its benefits for a small object localization task.

\vspace{5px}\noindent\textbf{Experimental setup.}
We compare our $TD$ proposal with its respective $BU$ baseline on MNIST, Fashion-MNIST and CIFAR10.
For the simpler MNIST tasks we consider as baselines the ``LeNetFC", a fully-convolutional variant of LeNet \cite{lecun1998gradient} and following \cite{lin2013network}, a lightweight version of the NIN (Network-In-Network) architecture, namely ``NIN-light" with reduced filters. The original architecture was used for the CIFAR10 task, along with the ResNet32 introduced in \cite{he2016deep} incorporating the pre-activation unit of \cite{he2016identity}. Batch Normalization \cite{ioffe2015batch} is used in all the networks prior to the non-linearities.
The corresponding $TD$ networks are defined based on their $BU$ baselines. 
Table \ref{tab:BUvTD_params} depicts the number of parameters of different models.
For $TD$ we consider three variants: 
$TD$ -- which is mirroring the $BU$ architecture also in terms of filter depth; $TD_{uni}$ using uniform filter depth; and $TD_{rev}$ where the filter depth of the $TD$ is reversed, thus following the filter depth of $BU$.
There is an increase in the number of parameters for the $TD$ networks, because we need additional convolutional layers for merging the high and low resolution information. 

We abide by the setup found in the initial publications for the $BU$ models. 
For the $TD$ networks we performed a linear search for learning rate, batch size, and weight decay. 
For all cases we train with a 90/10 train/val split, using SGD with momentum of 0.9 and a 3-stage learning rate decay scheme, dividing the learning rate by 10 at $50\%$ and $80\%$ of the total number of epochs. 
For the CIFAR10 dataset we test with and without augmentation---employing horizontal translations and flips. We repeat runs four times, with dataset reshuffling and extracting new training and validation splits, and report mean and standard deviation of the test accuracy. 
\begin{table}
	\begin{center}
	    \resizebox{0.4\textwidth}{!}{\begin{tabular}{lrrrr}
	        \toprule
			\multirow{2}{*}{Model} & \multicolumn{4}{c}{\#parameters}\\
			\cmidrule{2-5} \\
			{} & $BU$ & $TD$ & $TD_{uni}$  & $TD_{rev}$ \\
			\midrule
			LeNetFC   & 8k       & 14k      & 23k           & 58k           \\ 
			NIN-light & 62k      & 213k     & 215k          & 214k          \\ 
			ResNet32  & 468k     & 528k     & 320k          & 563k          \\
			NIN       & 970k     & 3,368k   & 3,397k        & 3,388k        \\ 
			\bottomrule
	    \end{tabular}}
	\end{center}
	\caption{Number of trainable parameters for the different architectures considered. Different rows correspond to the different baseline architectures and columns indicate the bottom-up model and the three top-down variants with different filter arrangements (section~\ref{sec:filters_arrangement}).
	There is an increase in the number of parameters for the $TD$ networks, because they merge the high and low resolution information using additional convolutional layers.}
	\label{tab:BUvTD_params}
	\vspace{-0.3cm}
\end{table}

\subsection{Exp. 1: Bottom-up versus top-down}\label{sec:BUvTD}
\begin{figure*}
    \begin{center}
        \includegraphics[width=\linewidth]{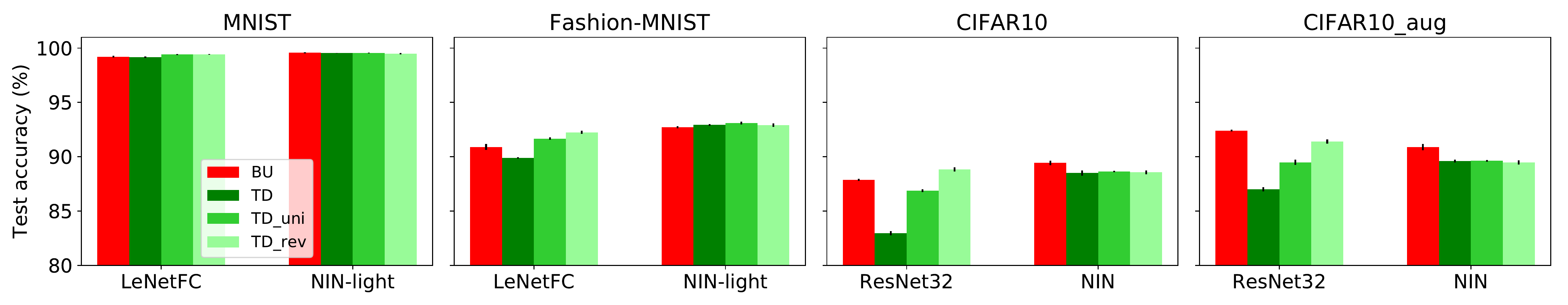}
    \end{center}
    \caption{\textbf{Exp 1:} Comparison of MNIST, Fashion-MNIST, CIFAR10, and CIFAR10\_aug (with augmentation) mean test accuracies between $BU$ and the three different configurations of $TD$ proposed in subsection \ref{sec:filters_arrangement}. 
    $TD$ networks perform on par with, and at times surpassing, the baseline performance of its respective $BU$. Regarding filter depth configurations, $TD_{rev}$ displays the highest performance, at the cost of increased parameters. Considering the small gap in performance and the increased cost for $TD_{rev}$, we henceforth adopt the $TD$ configuration.}
    \label{fig:BUvTD}
\end{figure*}
\begin{figure*}
    \begin{center}
	    \includegraphics[width=\linewidth]{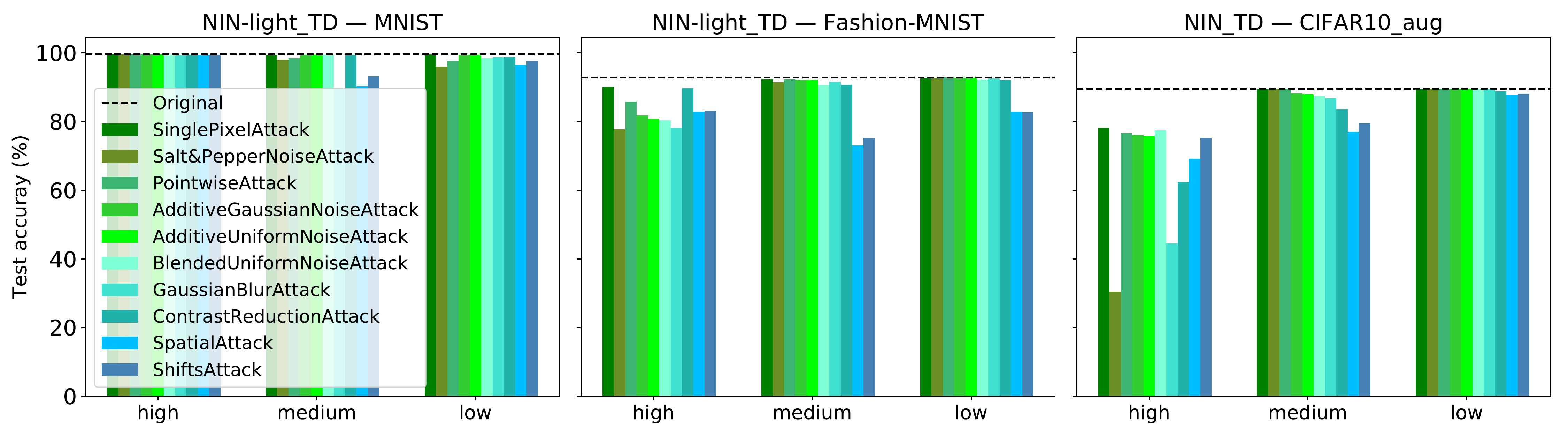}    
	\end{center}
	\caption{\textbf{Exp 2:} Test accuracy when extracted adversarial perturbations are fed to either the highest, medium, or lowest scale input of the $TD$ network (refer to figure \ref{fig:idea}), using the NIN-light baseline on MNIST and Fashion-MNIST, and NIN on CIFAR10. 
	The remaining two inputs are fed the original, unperturbed samples.
	As the dataset becomes more challenging, the highest vulnerability moves from the medium input to the highest scale input. 
	This is attributed to the absence of information in the high frequency region for the simpler cases: i.e. MNIST. (See the appendix for additional results.)}
	\label{fig:adversarial_inputs}
\end{figure*}

Figure \ref{fig:BUvTD} shows the test accuracy of the considered models across datasets.
The $TD$ networks are on par with, and in some cases surpassing the corresponding baseline $BU$ performance. 
When considering the different filter depth configurations, $TD_{rev}$ performs best due to increased representational power at higher scales, coming though at cost of increased complexity. The NIN architecture adopts a close to uniform filter arrangement, hence the three $TD$ variants reach roughly the same performance.
We adopt the $TD$ variants henceforth, on account of the small gap in performance and reduced complexity. 
This experiment provides empirical evidence of the applicability of the proposed pipeline to different network architectures.

\begin{figure*}
	\begin{center}
	    \includegraphics[width=\linewidth]{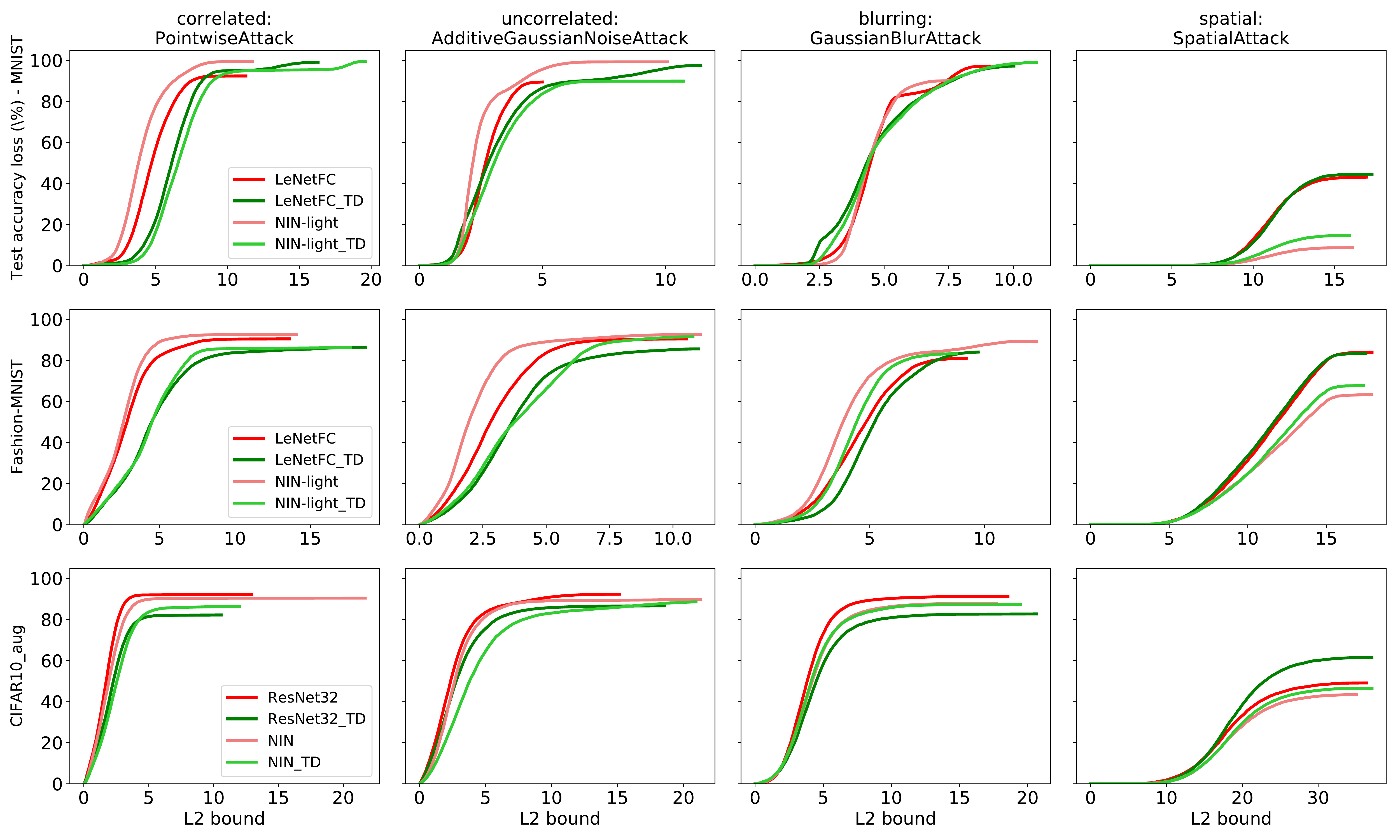}
	\end{center}
	\caption{\textbf{Exp 2:} Comparison of adversarial robustness considering different datasets, models and attacks. 
	The x-axis of each figure corresponds to the $L2$ distance between the original and the perturbed image and the y-axis is the introduced loss in test accuracy. 
	A lower curve suggests increased robustness. 
	Green curves corresponding to $TD$ are consistently underneath the respective red curves of the $BU$ networks, for most attacks. 
	The $TD$ networks are more robust against both correlated and uncorrelated noise attacks due to the coarse-to-fine processing, suppressing high frequency information on earlier stages. 
	Additionally, the blurred downsampling offers enhanced robustness against blurring attacks. For spatial attacks, we see no increased robustness.
	(See the appendix for additional results.)}
	\label{fig:adversarial_robustness}
\end{figure*}

\subsection{Exp. 2: Adversarial robustness}

We evaluate the robustness of $BU$ versus $TD$ against various attacks, where we attack the test set
of each dataset using the Foolbox \cite{rauber2017foolbox}. 
For all the attacks, the default parameters were used. 
To make the attack bound tighter, we repeat each attack three times and keep the worst case for each
to define the minimum required perturbation for fooling the network.

Figure~\ref{fig:adversarial_robustness} provides for each attack, plots of loss in test accuracy versus the $L2$ distance between the original and the perturbed input. 
$TD$ networks are visibly more resilient against attacks introducing uncorrelated noise, due to the coarse-to-fine processing adopted, with downscaled inputs diminishing the noise. 
For attacks introducing correlated noise such as the ``Pointwise" attack \cite{schott2018towards}, 
the perturbed pixels tend to lie in smooth regions of the image. 
Thus each single pixel value of 0 (or 1) in a region of 1s (or 0s) essentially acts as a Dirac delta function. Based on the convolutional nature of CNNs this type of attack ``pollutes" the input with imprints of the learned filters\footnote{For imperfect delta function, this yields blurred versions of the filters.}, which gradually span a greater part of the feature map as more convolutions are applied. 
Due to the highly correlated nature of the perturbation, the blurred downsampling can not completely eradicate the noise, but helps decrease the introduced pollution. 
On the contrary, for $BU$ networks, the noise is directly propagated down the network. 
Additionally, the blurred downsampling wired in the network architecture offers enhanced robustness against blurring attacks, as the network encounters the input image at multiple scales during training, and is, thus, more resilient to resolution changes.
Since anti-aliasing before downsampling is suggested to better preserve shift-invariance \cite{zhang2019making}, we expected our networks to also be more robust against the ``Spatial" attack \cite{engstrom2017exploring}. 
However, no enhanced robustness is reported for $TD$ networks; a substantial difference in robustness is observed for ResNet32, which could be due to the performance gap measured in \textbf{Exp 1} between the $TD$ and its $BU$ baseline. We also tested with the $TD_{uni}$ and $TD_{rev}$ variants of the ResNet32 architecture, with respective results provided in the appendix.

To get a better insight on $TD$ robustness, we introduce the generated attacks to a single resolution branch of the $TD$ networks using the NIN-light architecture on MNIST and Fashion-MNIST, and NIN on CIFAR10.
This is displayed in figure \ref{fig:adversarial_inputs}.
We feed the extracted perturbations to either the low, medium or high resolution input branch, as illustrated in the model architecture in figure~\ref{fig:idea}.
For the simpler MNIST task, the medium-resolution input of the network is the most vulnerable, which is mainly attributed to the absence of information in the high frequency region of the input's spectrum. Moving to more challenging Fashion-MNIST and CIFAR10 tasks, the high frequency input becomes the easiest path for fooling the network. 
Please see the appendix for additional results when perturbing two inputs simultaneously.

\subsection{\textbf{Exp 3:} Explainability and localization}
\begin{figure*}[h!]
    \begin{center}
        \includegraphics[width=\linewidth]{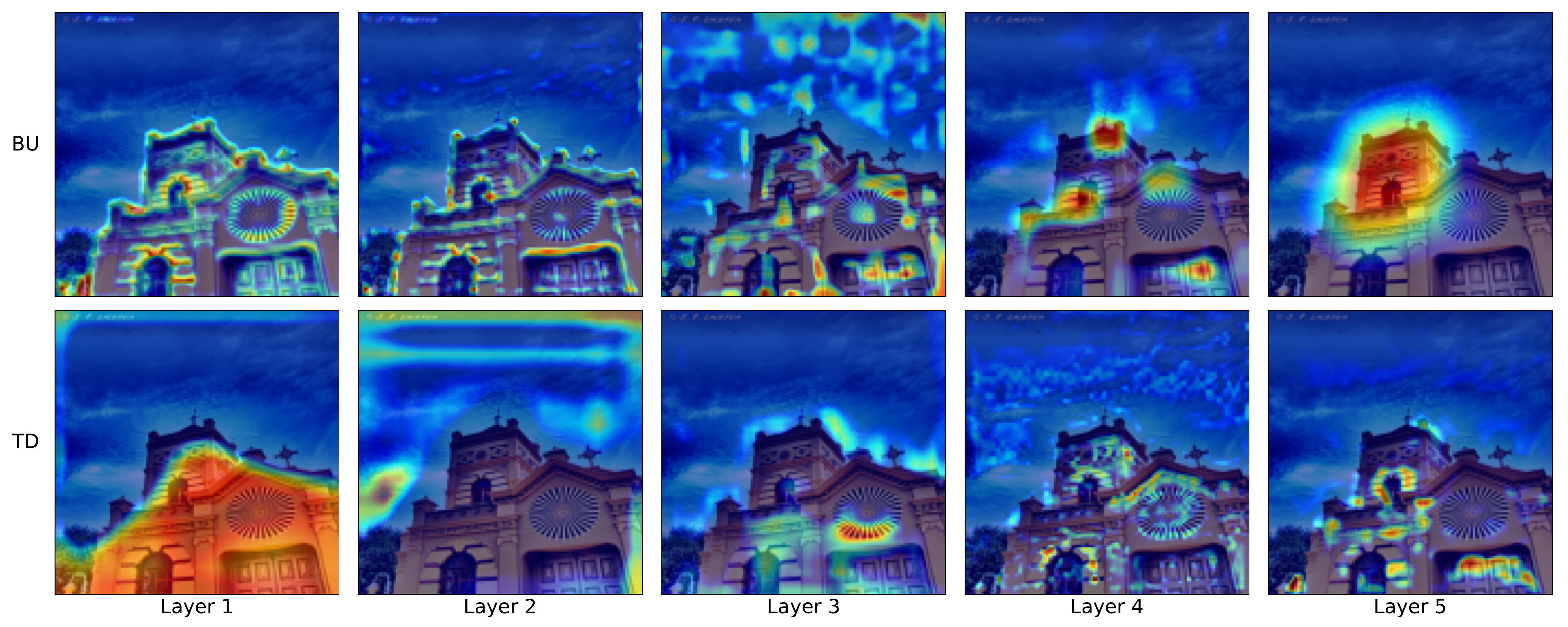}    
    \end{center}
    \caption{\textbf{Exp 3.(a):} Fine-to-coarse versus coarse-to-fine processing. 
    We show Grad-CAM heatmaps for ResNet18 $BU$ versus its respective $TD$, trained on the Imagenette dataset \cite{imagenette} for a random validation image. 
    Higher layer index means increased depth in the architecture: 
    ``Layer 1" corresponds to the activation of the input to the first group of residual blocks,
    and ``Layer 2" to ``Layer 5" to the activations of the output of each of these four groups, each one corresponding to different spatial resolution.
    \textbf{Top}: the $BU$ network, employing fine-to-coarse processing. 
    \textbf{Bottom}: the respective $TD$ network following the opposite path, starting with a holistic representation and gradually adding higher frequency information in deeper layers.
    }
    \label{fig:coarse_to_fine}
\end{figure*}
\begin{figure*}[h!]
    
    \begin{center}
        \includegraphics[width=\linewidth]{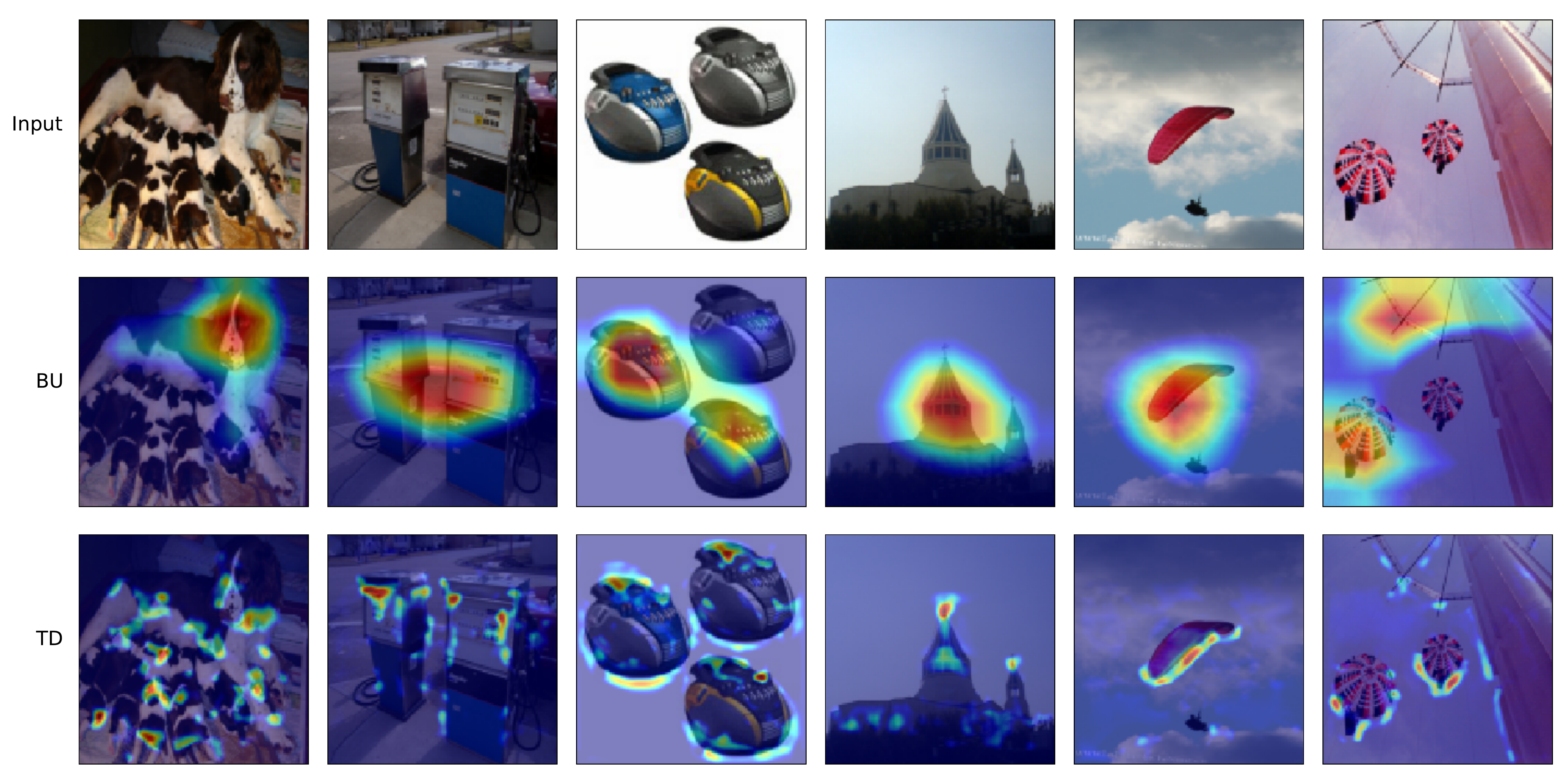}    
    \end{center}
    \caption{\textbf{Exp 3.(a):} Grad-CAM heatmaps corresponding to the last convolutional layer in the network.
    \textbf{Top}: The original input image, randomly selected from the validation set.
    \textbf{Middle}: Corresponding Grad-CAM heatmaps for the $BU$ ResNet18. 
    \textbf{Bottom}: Grad-CAM heatmaps for the $TD$ ResNet18. 
    Contrary to the coarse output of the $BU$, the $TD$ network outputs high frequency feature maps, based on which the final classification is performed. 
    $TD$ recognized objects based on their fine-grained attributes: such as the spots on the dogs,
    or the cross on the church, or shape information. (See the appendix for additional results.)}
    \label{fig:gradcam_heatmaps}
\end{figure*}

\begin{figure}[h!]
    \begin{center}
        \includegraphics[width=\linewidth]{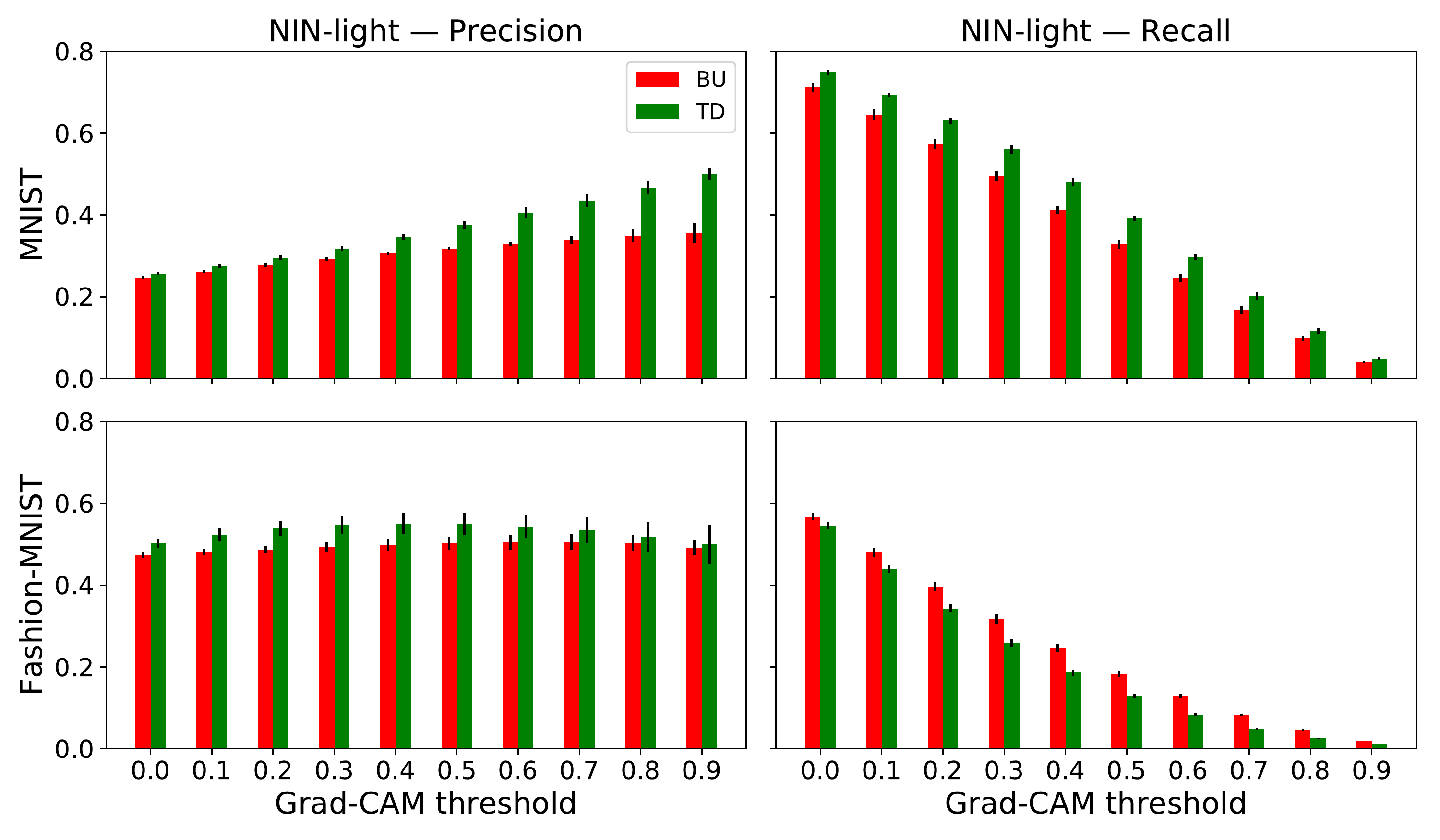}    
    \end{center}
    \caption{\textbf{Exp 3.(b):} Precision and recall for the MNIST and Fashion-MNIST datasets using the NIN-light architecture. 
    The numbers are reported over four runs and we also plot standard deviations.
    For each run, models are trained from scratch and the set of TP (true positive), FP (false positive), FN (false negative) is computed, between the Grad-CAM heatmaps and the segregated objects.
    The $TD$ model has higher precision on both MNIST and Fashion-MNIST due to more accurate object localization, while having slightly lower recall than $BU$ on the Fashion-MNIST.}
    \label{fig:object_log}
    \vspace{-0.3cm}
\end{figure}

\vspace{5px}\noindent\textbf{(a) Grad-CAM heatmap visualizations.}
Grad-CAM \cite{selvaraju2017grad} provides class-discriminative localization maps, based on the feature maps of a convolutional layer, highlighting the most informative features for the classification task. 
Here, we use the features of the last convolutional layer. 
The extracted heatmap is restored to the original image scale, thus producing a coarser map in the case of the $BU$ whose feature map size at the final layer is smaller. 
On the contrary, for $TD$ the corresponding scale of the feature maps matches the scale of the input, hence Grad-CAM outputs a finer map.

The Grad-CAM heatmaps corresponding to a $BU$ and $TD$ network are provided in figure \ref{fig:coarse_to_fine}.
These are obtained from various layers of a ResNet18 architecture \cite{he2016deep} trained on the Imagenette dataset \cite{imagenette}. 
For further information about the setup please refer to the appendix. 
``Layer 1" corresponds to the activation of the input to the first group of residual blocks,
and ``Layer 2" to ``Layer 5" to the activations of the output of each of these four groups, each one corresponding to different spatial resolution. 
The visualizations demonstrate that $TD$ follows an opposite, coarse-to-fine path starting from a coarser representation and gradually enriching it with higher frequency information. 
Hence, $TD$ networks do not only mirror the $BU$ solely in the architectural design, but also in their learning process.

Additional heatmaps corresponding to correctly classified images, taken from the last convolutional layer of the networks are visualized in figure \ref{fig:gradcam_heatmaps}. 
The figures depict the coarse localization in $BU$ versus the fine localization in $TD$. 
We selected intentionally images with multiple objects. 
The $TD$ networks recognize objects based on fine-grained information: such as the spots on the dog, the cross on the church or boundary information of various objects.


\vspace{5px}\noindent\textbf{(b) Weakly-supervised object localization.}
For a quantitative evaluation of the localization abilities of $TD$, we used the MNIST and Fashion-MNIST datasets and the NIN-light model as a backbone architecture. 
Figure \ref{fig:object_log} shows mean precision and recall scores for the $TD$ and $BU$ models over four runs. For each run models were trained from scratch, then TP (true positive), FP (false positive), FN (false negative) values were computed between the Grad-CAM heatmaps and the thresholded objects, corresponding to the test set of the considered task. 
We used a threshold empirically set to $t=0.2$. 
Based on the computed values precision and recall scores were extracted and aggregated over the four runs. For a fair comparison only the samples correctly classified from both $TD$ and $BU$ were considered.
The $TD$ models report higher precision for both tasks considered, suggesting finer object localization. The lower recall scores for the Fashion-MNIST is attributed to the higher number of FN compared to the $BU$ model. The larger object sizes of the Fashion-MNIST task, along with the coarse output of the $BU$ model, being able to capture a greater extent of them, leads to fewer FN. 
On the contrary, the $TD$ models focus on finer aspects of the objects, which are informative for the classification task.
Considering the fine-grained focus in the Grad-CAM outputs and the potential for weakly-supervised object localization, $TD$ networks comprise a promising direction for future research. 


\section{Discussion}
\label{sec:discussion}
The current work aims at providing a fresh perspective on the architecture of CNNs, which is currently taken for granted.  
The coarse-to-fine pathway is biologically inspired by how humans perceive visual information: first understanding the context and then filling in the salient details. 

One downside of our proposed $TD$ networks is that expanding dimensions at increased network depth leads to memory and computational bottlenecks.
This is due to the feature map size being larger at higher depths.
Moreover, for the same reason, adding fully-connected layers before the output layer of the $TD$ architectures leads to a vast increase in the number of model parameters. 
Hence, fully convolutional networks are preferable. 
This increase in memory is also more visible with large-scale datasets such as ImageNet~\cite{deng2009imagenet}. 
A simple workaround requiring no architectural adaptations would be to employ mixed-precision training, which would decrease the memory requirements, but would increase the computational complexity. 
Instead of increasing the spatial resolution of the feature maps at later depths, we could use patches of the input of limited sizes. 
The selection of these informative patches could be defined using the Grad-CAM heatmaps by selecting the high-activation areas of the heatmap, or considering self-attention mechanisms \cite{xu2015show}.
In addition to addressing the aforementioned limitations, we find the weakly-supervised setting to be a promising area of future research.

\section{Conclusion}\label{sec:conclusion}
In the current work, we revisit the architecture of conventional CNNs, aiming at diverging from the manner in which resolution is typically processed in deep networks.
We propose novel network architectures which reverse the resolution processing of standard CNNs.
The proposed $TD$ paradigm adopts a coarse-to-fine information processing pathway, starting from the 
low resolution information, providing the visual context, and subsequently adding back the high
frequency information. 
We empirically demonstrate the applicability of our proposed $TD$ architectures when starting from 
a range of baseline architectures, and considering multiple visual recognition tasks. 
$TD$ networks exhibit enhanced robustness against certain types of adversarial attacks.
This resistance to adversarial attacks is induced directly by the network design choices. 
Additionally, the high spatial dimensions of the feature maps in the last layer significantly enhance the explainability of the model, and demonstrate potential for weakly-supervised object localization tasks.

{\small
\bibliographystyle{ieee_fullname}
\bibliography{bibliography}
}

\newpage
\appendix
\section*{\Large\textbf{Appendix}}
\vspace{0.3cm}
\section{\textbf{Exp 2:} Adversarial robustness}
\label{sec:adversarial_robustness_appendix}

\begin{figure*}
    \centering
    \includegraphics[width=0.8\linewidth]{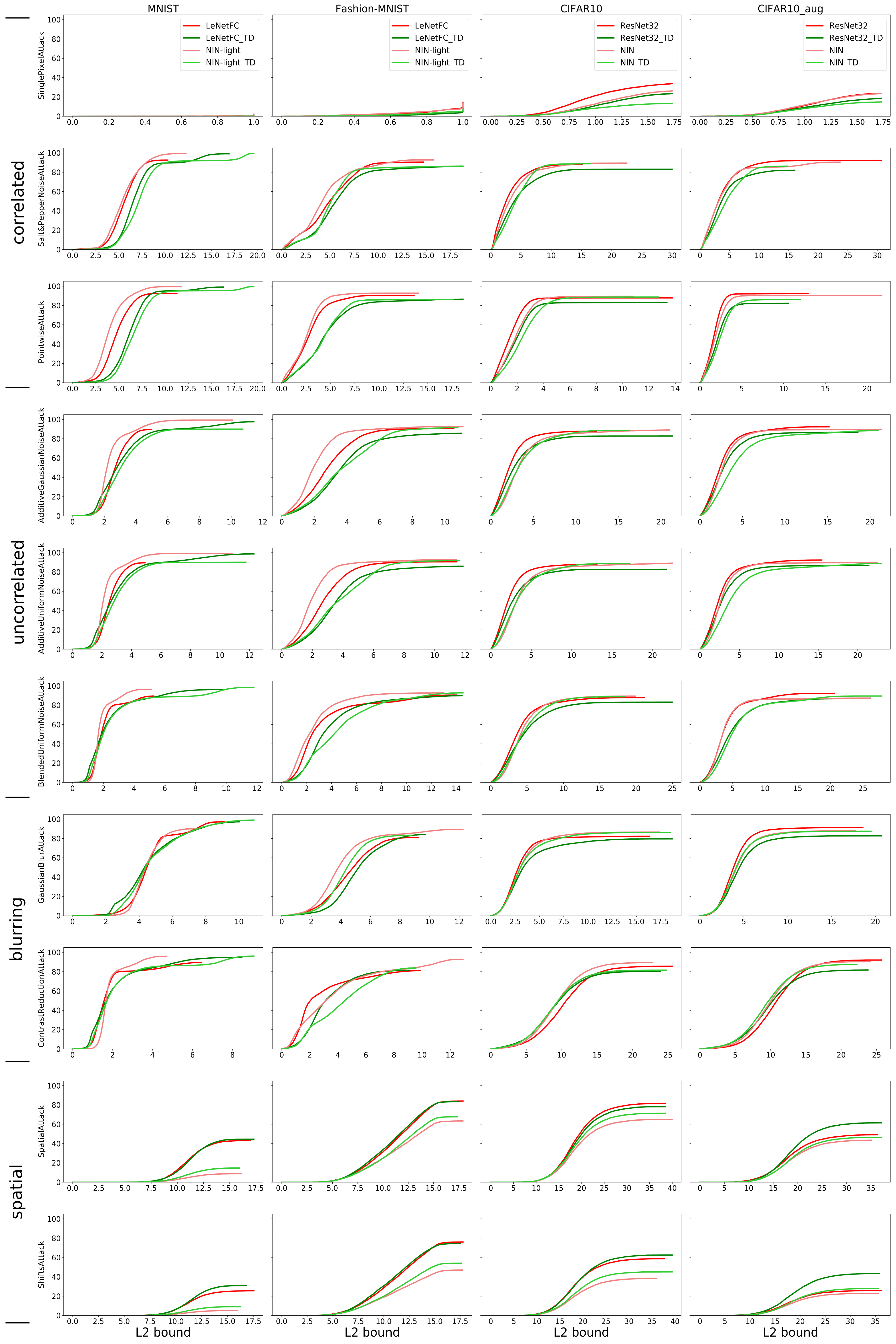}
    \caption{\textbf{Exp:2} Complete set of results for the second experiment. Plots of test accuracy loss versus the $L2$ distance between original and perturbed input, where each column corresponds to a different task. $TD$ networks exhibit enhanced robustness against correlated/uncorrelated noise and blurring attacks.}
    \label{fig:adversarial_robustness_full}
\end{figure*}

\begin{figure*}
    \centering
    \includegraphics[width=1\linewidth]{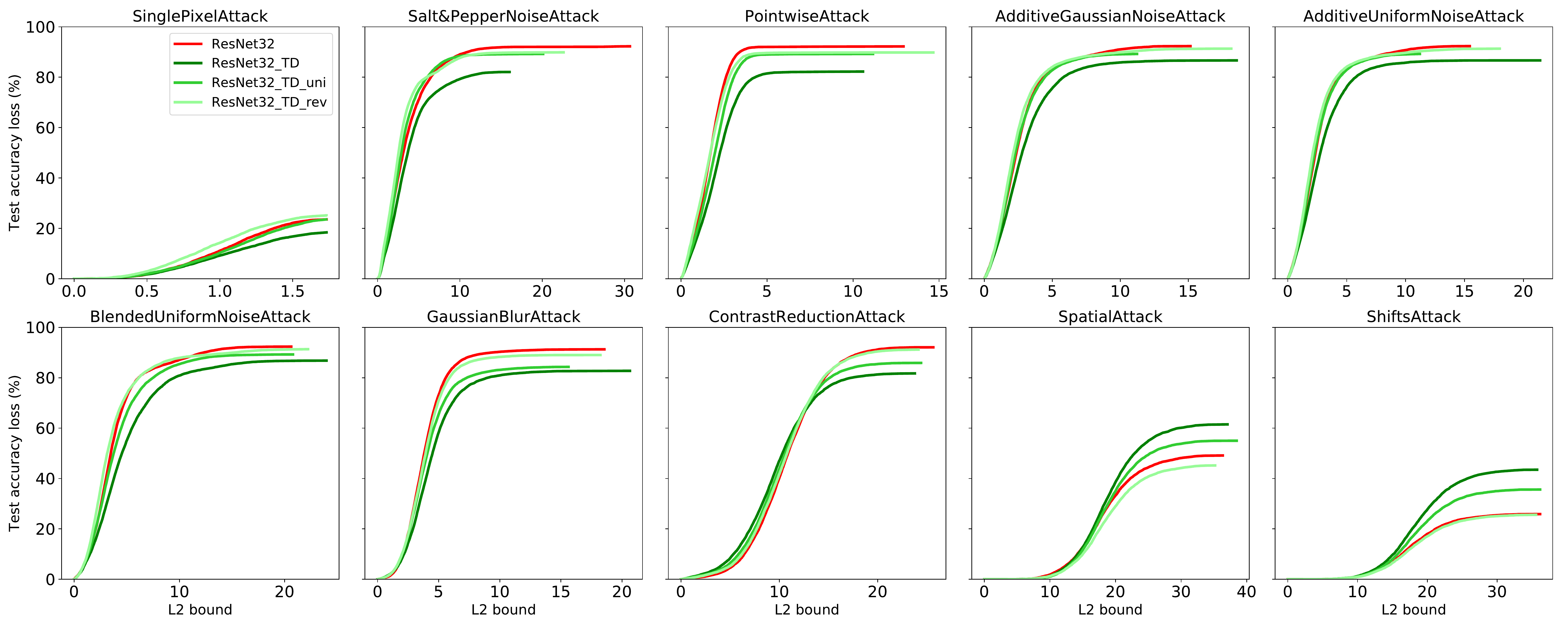}
    \caption{\textbf{Exp 2:} Test accuracy loss versus the $L2$ distance between original and perturbed input, for the CIFAR10-augmented and the ResNet32 architectures. 
    Robustness is enhanced for the spatial attacks, but in general $TD_{uni}$ and $TD_{rev}$ variants exhibit similar behaviour to the $BU$ baseline, which can be attributed to the increased filters at deeper layers.}
    \label{fig:adversarial_robustness_cifar_aug}
\end{figure*}

The entire set of results for the adversarial robustness experiment is provided in figure \ref{fig:adversarial_robustness_full}. ``ShiftsAttack" is a variant of Spatial attack \cite{engstrom2017exploring}, introducing only spatial shifts. $TD$ networks exhibit enhanced robustness against attacks introducing correlated/uncorrelated noise, as well as against blurring attacks.

Figure \ref{fig:adversarial_robustness_cifar_aug} presents the robustness results for the CIFAR10-augmented and the ResNet32 architecture variants. 
Clearly, $TD_{uni}$ and $TD_{rev}$ variants exhibit enhanced robustness against spatial attacks, however, they also have similar to the $BU$ behaviour against other attacks. 
This can be attributed to the increased number of filters at greater depth of the network, or equivalently increased scale of feature maps, thus greater contribution of the finer scales to the final output. However, finer scales are much more vulnerable against attacks. All in all, the reversal of the $BU$ network for the extraction of the $TD$ variant is not solely efficiency driven, keeping a roughly fixed computational complexity across layers, but also contributes to the network's robustness as well. Finally, we need to mention that the respective figure for the non-augmented CIFAR10 case tells the same story.

\begin{figure*}
\centering
    \includegraphics[width=\linewidth]{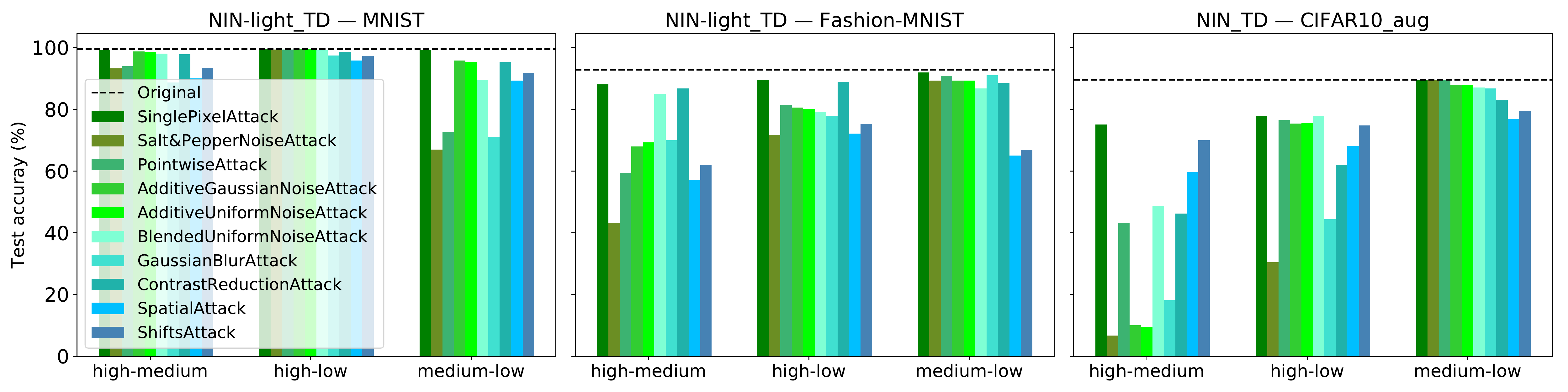}
    \caption{\textbf{Exp 2:} Reintroducing perturbations to two of the inputs of $TD$ model 
    when using a NIN-light backbone for MNIST and Fashion-MNIST, and the NIN backbone for CIFAR10. Clearly, perturbing the two highest scale inputs, ``high-medium" has the highest impact. Regarding the case of the simpler MNIST and the information gathered in the low to mid frequency region, the medium and the lowest scale input have the highest impact instead.}
    \label{fig:inputs_robustness_2}
\end{figure*}

Next, figure \ref{fig:inputs_robustness_2} presents the respective results for reintroducing the perturbation to two of the inputs of the network. Clearly, the highest and medium scale inputs are the most vulnerable ones, except for the simpler case of the MNIST dataset. The absence or scarce information in the high frequency region, yields the medium and lowest scale inputs as the ones with the highest impact.


\section{\textbf{Exp 3.(a):} Explainability}
\subsection{Imagenette training}
\label{sec:imagenette_training}
\begin{figure*}
    \centering
    \includegraphics[width=\textwidth]{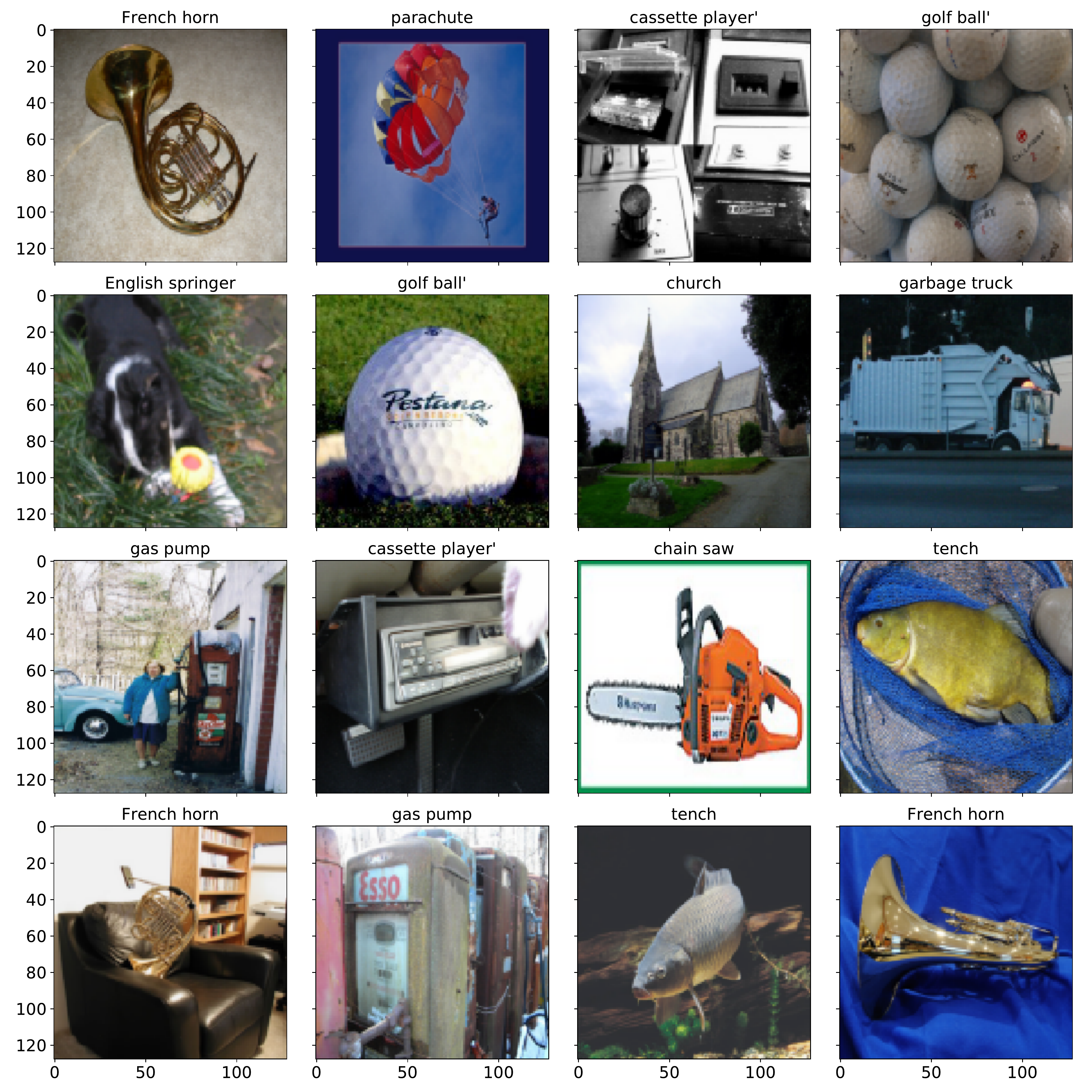}
    \caption{\textbf{Exp 3.(a):} Validation samples from the Imagenette dataset \cite{imagenette}, along with their corresponding ground truth labels. Samples are resized to $128\times128$.}
    \label{fig:imagenette}
\end{figure*}

Imagenette \cite{imagenette} is a 10-class sub-problem of ImageNet \cite{deng2009imagenet}, allowing experimentation with a more realistic task, without the high training times and computational costs required for training on a large scale dataset. A set of examples, along with their corresponding labels are provided in figure \ref{fig:imagenette}. The datasets contains a total of 9469, 3925 training and validation samples respectively. Training samples were resized to $156 \times 156$, from where random $128 \times 128$ crops were extracted; validation samples were resized to $128\times128$.

We utilized a lighter version\footnote{dividing the filters of the original architecture by 2.} of the ResNet18 architecture introduced in \cite{he2016deep} for Imagenette training, as this is a 10-class sub-problem, incorporating the pre-activation unit of \cite{he2016identity}. Additionally, the stride $s$ and the kernel extent $k$ of the first convolution for depth initialization were set to $s=1$ and $k=3$ respectively. Regarding training, a $128\times 128$ crop is extracted from the original image, or its horizontal flip, while subtracting the per-pixel mean \cite{krizhevsky2012imagenet}; the color augmentation of \cite{krizhevsky2012imagenet} is also used. For the $BU$ network a batch size of 128 is used and the network is trained for a total of 50 epochs with a starting learning rate of 0.1. As for the $TD$, increased memory footprint led to the reduction of the batch size to 64 and the adaptation of the starting learning rate and the total epochs to $0.05$ and 80. We trained with SGD with momentum of 0.9 and a weight decay of 0.001; we also adopted a 3-stage learning rate decay scheme, where the learning rate is divided by 10 at $50\%$ and $80\%$ of the total epochs. Regarding performance, $BU$ outperformed the $TD$ variant by roughly $4\%$. Grad-CAM is finally utilized for generating class-discriminate localization maps of the most informative features.

\subsection{Grad-CAM heatmap visualizations}
\begin{figure*}
    \centering
    \includegraphics[width=\linewidth]{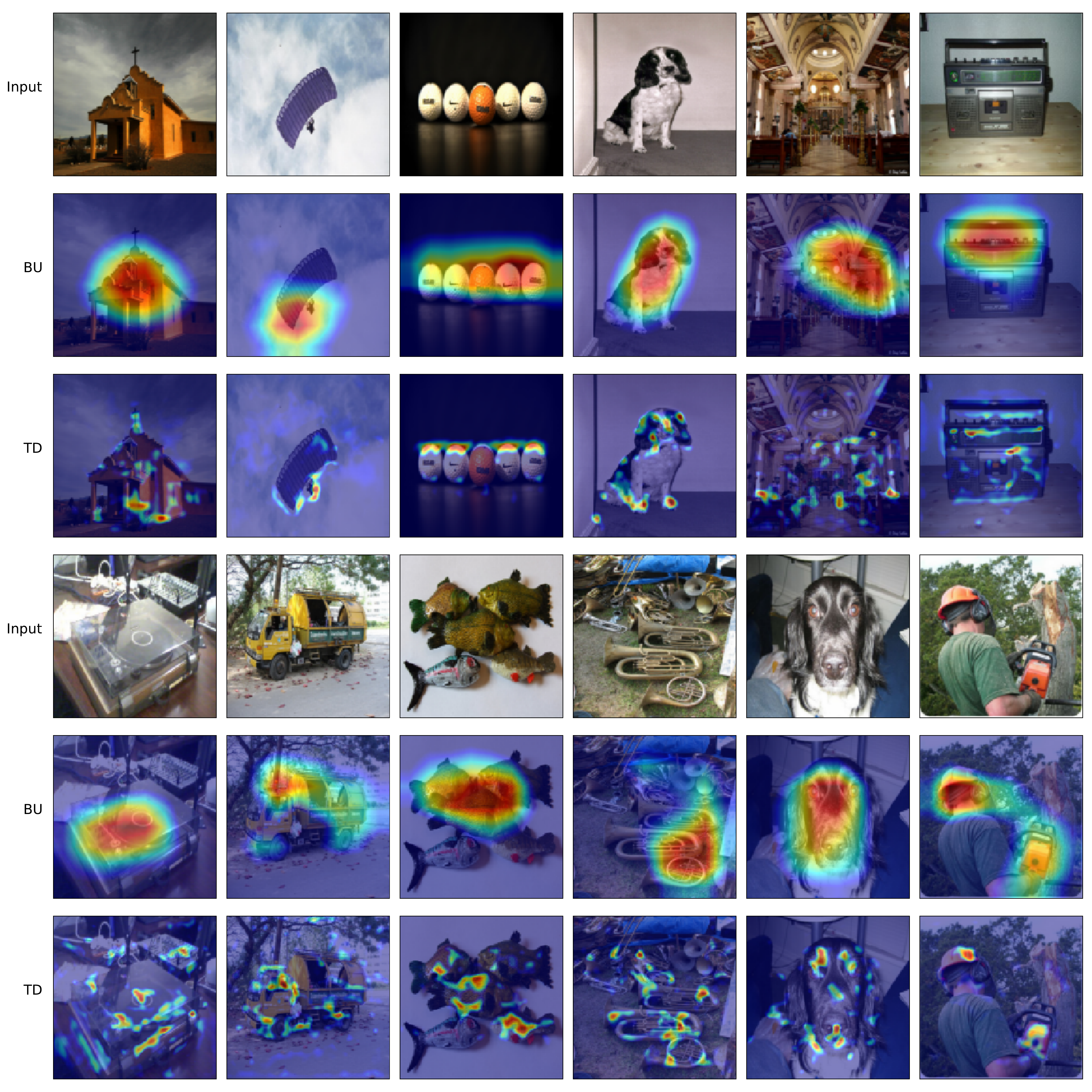}
    \caption{\textbf{Exp 3.(a):} Grad-CAM heatmaps visualization on validation images from the Imagenette dataset \cite{imagenette}, using a ResNet18 architecture for $BU$ and its corresponding $TD$ variant.
    All images are correctly classified.
    Rows 1 and 4 show the original Imagenette images; rows 2 and 5 show the $BU$ heatmaps, while 
    rows 3 and 6 visualize the $TD$ heatmaps.
    Focusing on local information rather than global information, may help the $TD$ to 
    be more precise for object detection than the $BU$ model.}
    \label{fig:gradcam_2}
\end{figure*}

Figure~\ref{fig:gradcam_2} displays some additional Grad-CAM visualizations.
The visualizations are obtained by using a ResNet18 architecture for the $BU$ networks and
its corresponding $TD$ variant. 
The original images are taken from the Imagenette dataset \cite{imagenette}.

The $TD$ model provides localized activations, focusing on certain informative aspects of the image, while the $BU$ model focuses on large connected areas.
Because of this difference we believe the $TD$ model may be more precise than the $BU$ model for 
tasks such as weakly-supervised object detection. 

\end{document}